\documentclass[runningheads]{llncs}

\usepackage{arxiv}

\usepackage{eccvabbrv}

\usepackage{graphicx}
\usepackage{booktabs}

\usepackage[accsupp]{axessibility}  

\usepackage{hyperref}

\usepackage{orcidlink}

\begin{document}

\title{Last-Layer-Centric Feature Recombination: Unleashing 3D Geometric Knowledge in DINOv3 for Monocular Depth Estimation} 

\titlerunning{Abbreviated paper title}

\author{Gongshu Wang\inst{1}\orcidlink{0000-0002-3984-058X}	\\
Aerospace Information Research Institute\\
	Chinese Academy of Sciences\\
	Beijing, China \\
	\texttt{wanggs@aircas.ac.cn} \\ 
\and
Zhirui Wang\inst{1} \\
Aerospace Information Research Institute\\
	Chinese Academy of Sciences\\
	Beijing, China \\
	\texttt{wangzr@aircas.ac.cn} \\ 
\and
Kan Yang\inst{1} \\
Aerospace Information Research Institute\\
	Chinese Academy of Sciences\\
	Beijing, China \\
    \texttt{yangkan@aircas.ac.cn} \\}
\authorrunning{F.~Author et al.}

\institute{Princeton University, Princeton NJ 08544, USA \and
Springer Heidelberg, Tiergartenstr.~17, 69121 Heidelberg, Germany
\email{lncs@springer.com}\\
\url{http://www.springer.com/gp/computer-science/lncs} \and
ABC Institute, Rupert-Karls-University Heidelberg, Heidelberg, Germany\\
\email{\{abc,lncs\}@uni-heidelberg.de}}

\maketitle

\begin{abstract}
Monocular depth estimation (MDE) is a fundamental yet inherently ill-posed task. Recent vision foundation models (VFMs), particularly DINO-based transformers, have significantly improved accuracy and generalization for dense prediction. Prior works generally follow a unified paradigm: sampling a fixed set of intermediate transformer layers at uniform intervals to build multi-scale features. This common practice implicitly assumes that geometric information is uniformly distributed across layers, which may underutilize the structural 3D cues encoded in VFMs.
In this study, we present a systematic layer-wise analysis of DINOv3, revealing that 3D information is distributed non-uniformly: deeper layers exhibit stronger depth predictability and better capture inter-sample geometric variation. Motivated by this, we introduce a Last-Layer-Centric Feature Recombination (LFR) module to enhance geometric expressiveness. LFR treats the final layer as a geometric anchor and adaptively selects complementary intermediate layers according to a minimal-similarity criterion. Selected features are fused with the last-layer representation via compact linear adapters.Extensive experiments show that LFR module consistently improves MDE accuracy and achieves state-of-the-art performance. Our analysis sheds light on how geometric knowledge is organized within VFMs and offers an efficient strategy for unlocking their potential in dense 3D tasks.

  \keywords{Monocular depth estimation \and 3D understanding \and Foundation model}
\end{abstract}

\section{Introduction}
\label{sec:intro}

Monocular Depth Estimation (MDE) aims to predict pixel-wise scene depth from a single RGB image. It is a critical task in computer vision with diverse applications, including autonomous driving\cite{M3D}, virtual reality\cite{vr}, embodied navigation\cite{Vla}\cite{3d}, and 3D reconstruction\cite{Zpressor}\cite{volsplat}. However, MDE is inherently ill-posed, as a 2D image can correspond to infinitely many possible 3D scenes. Early approaches sought to mitigate this ambiguity by incorporating handcrafted geometric priors\cite{DORN}\cite{adabins}\cite{Gedepth}\cite{Localbin} to guide model learning. Despite some success, these priors are often limited in their ability to generalize across diverse real-world scenarios\cite{idisc}.

The emergence of large-scale vision foundation models(VFMs) has created new opportunities for MDE. Pre-trained on massive datasets with proxy objectives, these models capture generic representations that can be effectively transferred to downstream tasks. Among them, DINO-style\cite{dino}\cite{DINOv2}\cite{dinov3} vision transformers (ViTs)\cite{vit} have shown excellent performance for both semantic and dense prediction tasks, particularly those involving 3D understanding. Notably, recent state-of-the-art (SOTA) models such as Depth Anything\cite{DA1}\cite{DA2} and VGGT\cite{Vggt} build upon DINO-family, leveraging its representations through fine-tuning on large-scale 3D datasets. While these approaches focus on data driven and training objectives, their architectural design largely follows a common paradigm: features from a fixed set of intermediate ViT layers are fed into a DPT\cite{DPT} decoder to produce multi-scale representations for final prediction.

Although this paradigm has proven effective, it may not fully exploit the rich geometric knowledge embedded in VFMs. Due to the weak inductive bias of ViTs\cite{park}, feature distributions vary significantly across layers and training objectives. For instance, attention maps in CLIP\cite{CLIP} differ markedly from those in DINO\cite{DINOv2}. Thus, understanding the internal distribution of 3D knowledge within DINO is crucial for designing more effective transfer strategies.

\begin{figure}[tb]
  \centering
  \includegraphics[height=6cm]{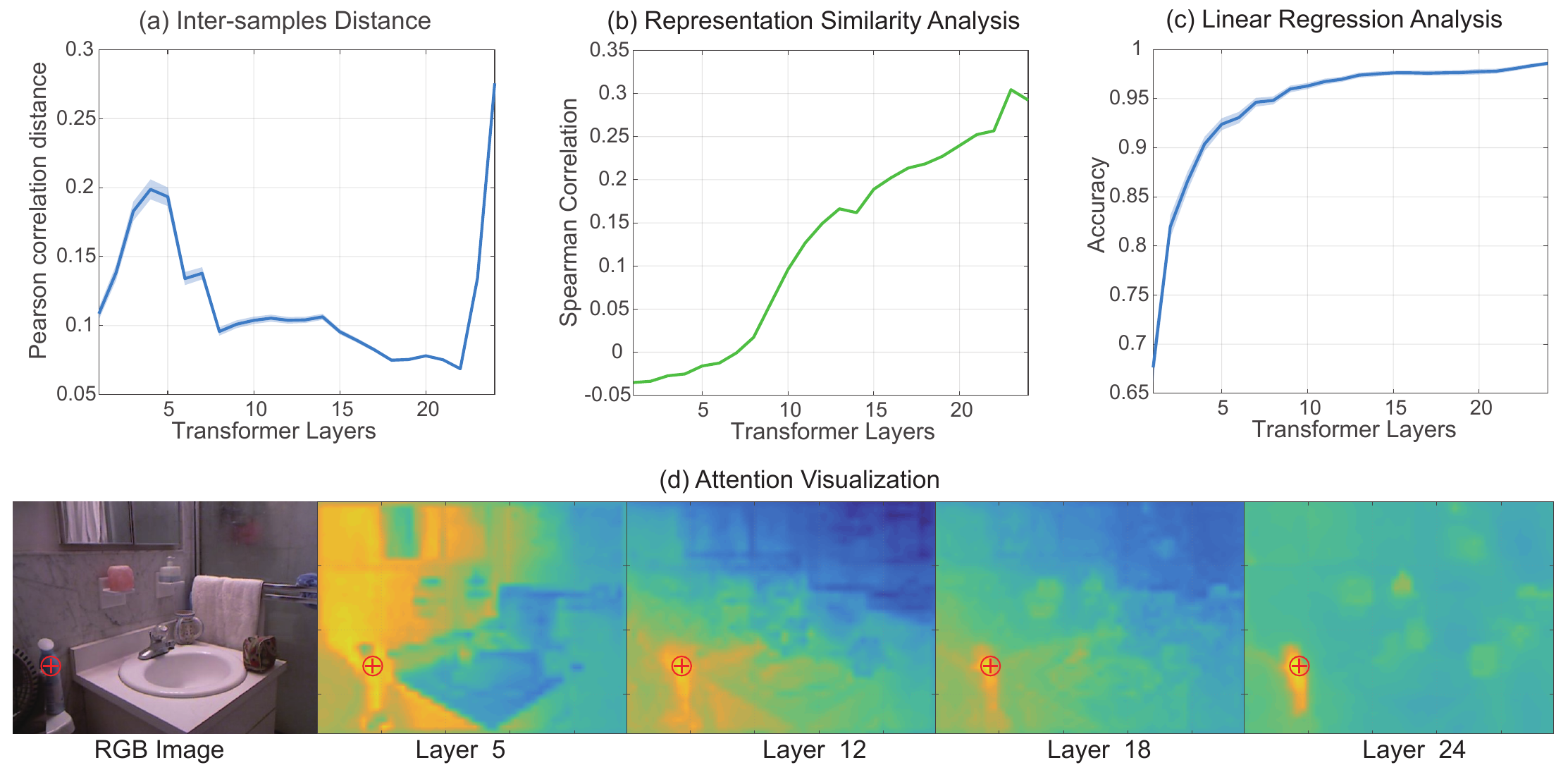}
  \caption{Layer-wise feature analysis of DINOv3-L.
For each transformer layer, we compute (a) inter-sample representational distance (mean pairwise Pearson correlation distance), (b) representational similarity to ground-truth depth (Spearman correlation between RDM and DDM), (c) depth predictability via linear regression. Shaded regions in (a) and (c) indicate 95\% confidence intervals. (d) visualizes cosine similarity between the anchor token and all other tokens (brighter indicates higher similarity).
  }
  \label{fig:attn}
\end{figure}

In this work, we first perform a thorough layer-wise statistical analysis of DINOv3-L\cite{dinov3} features(Fig. \ref{fig:attn}). We examine inter-sample representational distances, linear depth predictability, and representational similarity to ground-truth depth. Our analysis reveals that deeper layers encode richer and more explicit 3D information, and are better at capturing variations in 3D geometry across samples. In addition, prior works have shown that last-layer features already contain sufficient information to describe multi-scale representations\cite{PLAIN}, and that fusing last-layer with intermediate features benefits dense prediction\cite{Resclip}.Together, these insights suggests that deeper features should play a more central role in MDE.

\begin{figure}[tb]
  \centering
  \includegraphics[height=6cm]{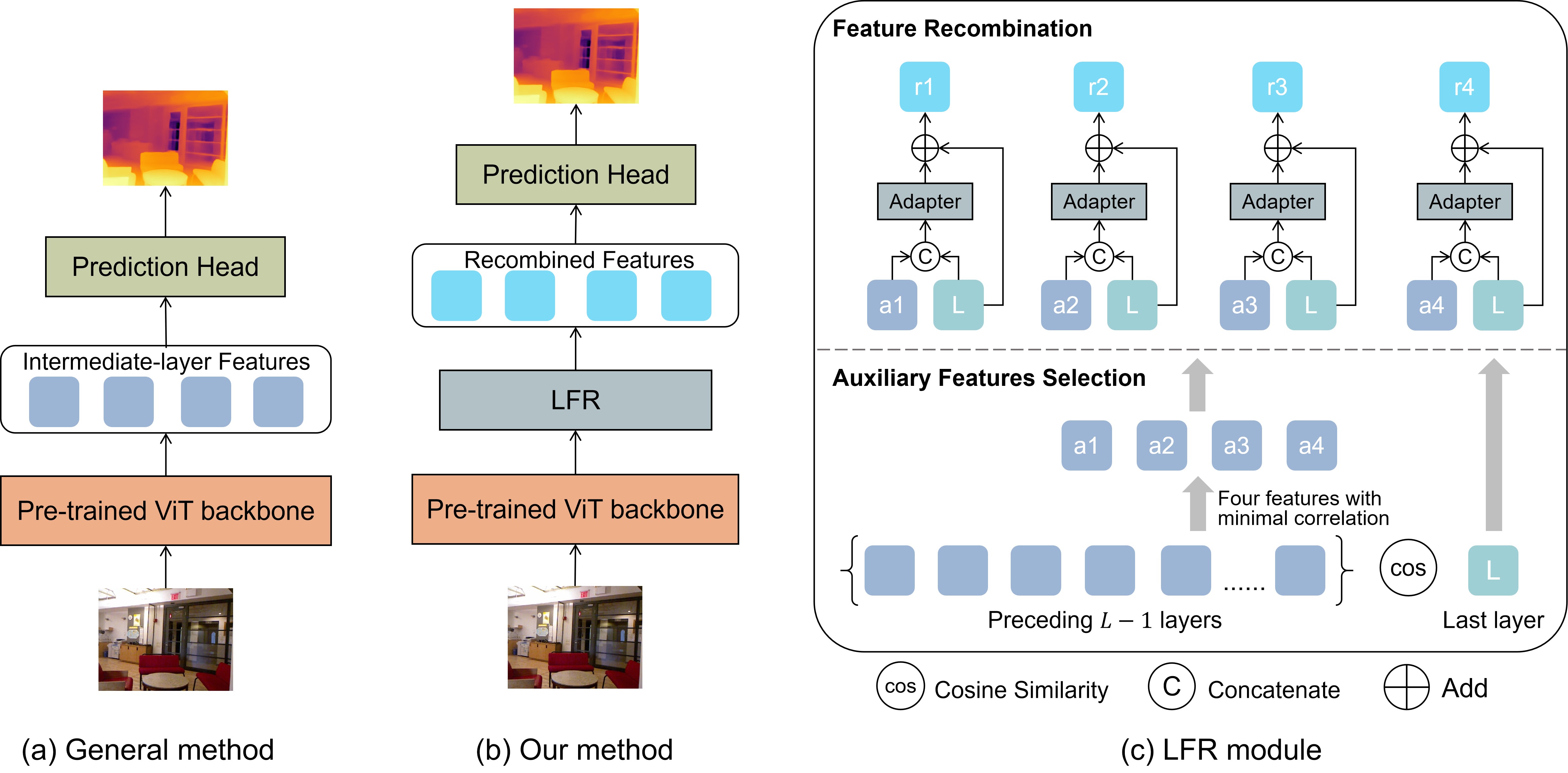}
  \caption{Overview of the proposed method. (a) Generic dense prediction pipeline based on a ViT backbone. (b) Our improved architecture with the Last-layer-centric Feature Recombination (LFR) module inserted between the backbone and the prediction head. (c) Detailed illustration of the LFR module: the last-layer features serve as the dominant representation; four complementary intermediate layers are selected via a minimal-similarity criterion; lightweight linear adapters fuse each selected auxiliary feature with the last-layer features to produce recomposed features for the decoder.
  }
  \label{fig:method}
\end{figure}

Motivated by these findings, we propose a simple yet effective last-layer-centric feature recombination (LFR) module (Fig. \ref{fig:method}). This plug-and-play module sits between the ViT backbone and the DPT decoder. It regards the last-layer features as the dominant geometric anchor and adaptively selects a small set of intermediate layers that are maximally complementary to this anchor. Concretely, we introduce a minimal-similarity-based selector that picks layers whose features have the lowest similarity to the last-layer features, maximizing knowledge complementarity. A lightweight adapter—composed of simple linear layers—then fuses each selected auxiliary feature with the last-layer features to produce recomposed representations. The final depth prediction is obtained as a weighted sum of per-level predictions. Our experiments on standard MDE benchmarks show that LFR consistently improves DINOv3-based MDE and attains new SOTA accuracy with minimal overhead.We summarize our contributions as follows:

(1) We present the first systematic layer-wise analysis of DINOv3 for 3D geometry, demonstrating that deeper layers contain denser and more explicit geometric information that is highly predictive for depth.

(2) We propose LFR, a simple and effective last-layer-centric recombination module that adaptively selects and fuses complementary intermediate features to amplify the geometric signal for MDE.

(3) We validate LFR on multiple benchmarks, showing consistent gains over strong baselines and achieving SOTA results with efficient computation and strong zero-shot generalization.

\section{Related Work}

\subsection{Monocular Depth Estimation}
MDE aims to predict per-pixel depth from a single RGB image and has been advanced substantially by deep learning. The literature can be grouped into three main directions:

\textbf{(1) Geometric Priors.} Early methods sought to reduce scale ambiguity by incorporating explicit geometric constraints derived from human knowledge of the visual world. For instance, piecewise planarity priors decompose scenes into planar segments\cite{single}\cite{Planenet}, while normal consistency enforce alignment between normals derived from predicted and ground-truth depth\cite{Geonet}\cite{surface}. Although effective in constrained settings, these handcrafted priors often fail to generalize across diverse, unconstrained environments and may inadvertently restrict model expressivity.

\textbf{(2) Generative Models.} Motivated by the success of diffusion models in image synthesis, several approaches formulate MDE as a conditional generation task. DDP\cite{DDP} adopts a noise-to-depth paradigm, conditioning the diffusion process on the input image. VPD\cite{VPD} leverages denoising UNets as backbones to extract multi-scale features. ECoDepth\cite{Ecodepth} employs a conditional diffusion architecture with the semantic context. DAR\cite{dar} draws inspiration from VAR\cite{var} and recasts MDE as a scale-wise autoregressive prediction task. These methods can achieve strong performance but typically require expensive iterative inference and large computational costs.

\textbf{(3) Large-Scale Data-Driven Methods.} The introduction of MiDaS\cite{MiDas} marked a paradigm shift by demonstrating that training on diverse, mixed datasets can yield strong zero-shot generalization. Building on this, ZoeDepth\cite{Zoedepth} and Depth Anything\cite{DA1}\cite{DA2} further scale up data—leveraging 62M unlabeled images via self-supervised learning—to achieve remarkable zero-shot capabilities. However, such approaches demand extensive data collection and training resources.

\subsection{Vision Foundation Models}
The emergence of VFMs pre-trained on massive datasets has reshaped the dense prediction tasks. Prominent examples include CLIP\cite{CLIP}, the SAM family\cite{sam}\cite{sam2}\cite{sam3}, and the DINO family\cite{dino}\cite{DINOv2}\cite{dinov3}. Among these, DINO—a purely vision-based VFM built upon the ViT\cite{vit} architecture—has demonstrated exceptional performance on both semantic tasks and dense prediction tasks.

Notably, a growing work in 3D understanding leverages the dense features of DINO models. For instance, MoGe\cite{Moge}, a SOTA open-domain monocular geometry model; VGGT\cite{Vggt}, which infers key 3D attributes from single or multiple views; and the Depth Anything series\cite{DA1}\cite{DA2}, which achieves SOTA zero-shot MDE performance—all benefit substantially from DINO pre-trained weights. With the recent release of DINOv3\cite{dinov3}—a 7-billion-parameter model trained on approximately 17 billion images—yield notably improved geometric encodings. Remarkably, simply transferring DINOv3 to relative depth estimation already surpasses the Depth Anything series. Anydepth pushes this further by attaching a lightweight prediction head to DINOv3, achieving superior results in relative depth estimation\cite{AnyDepth}. These findings underscore the rich 3D geometric knowledge encoded in DINOv3 and highlight its untapped potential.

Our work departs from prior practice by explicitly studying how 3D knowledge is arranged across DINOv3 layers and by proposing a principled, last-layer-centric recombination scheme that leverages this structure for MDE.
 
\begin{table}[tb]
  \caption{Comparison on NYU Depth v2. † denotes generative models that require multiple forward passes. '*' denotes data-driven methods that require large-scale pre-training on relative depth. Bold indicates the best result; underline indicates the second best.
  }
  \label{tab:nyu}
  \centering
  \setlength{\tabcolsep}{3pt}
  \begin{tabular}{@{}lllllllll@{}}
    \toprule
    Method & Size & AbsRel↓ & RMSE↓ & log10↓ & SqRel↓ & \(\delta_1\)↑ & \(\delta_2\)↑ & \(\delta_3\)↑\\
    \midrule
    Eigenetal.\cite{Eigen} & 45M & 0.158 & 0.641 & - & - & 0.769 & 0.95 & 0.988\\
    DORN\cite{DORN} & 45M & 0.115 & 0.509 & 0.051 & - & 0.828 & 0.965 & 0.992\\
    BTS\cite{BTS} & 20M & 0.11 & 0.392 & 0.047 & 0.066 & 0.885 & 0.978 & 0.994\\
    AdaBins\cite{adabins} & 40M & 0.103 & 0.364 & 0.044 & - & 0.903 & 0.984 & 0.997\\
    DPT\cite{DPT} & 343M & 0.11 & 0.357 & 0.045 & - & 0.904 & 0.988 & 0.998\\
    P3Depth\cite{P3depth} & 43M & 0.104 & 0.356 & 0.043 & - & 0.898 & 0.981 & 0.996\\
    NeWCRFs\cite{NeWCRF} & 	270M & 0.095 & 0.334 & 0.041 & 0.045 & 0.922 & 0.992 & 0.998\\
    SwinV2-L\cite{swinv2} & 197M & 0.112 & 0.381 & 0.051 & - & 0.886 & 0.984 & 0.997\\
    Localbins\cite{Localbin} & 74M & 0.099 & 0.357 & 0.042 & - & 0.907 & 0.987 & 0.998\\
    PixelFormer\cite{PixelFormer} & 365M & 0.09 & 0.322 & 0.039 & 0.043 & 0.929 & 0.991 & 0.998\\
    MIM\cite{MIM} & 197M & 0.083 & 0.287 & 0.035 & - & 0.949 & 0.994 & 0.999\\
    \bottomrule
    VPD\cite{VPD} † & 600M & 0.069 & 0.254 & 0.030 & 0.027 & 0.964 & 0.995 & 0.999\\
    EcoDepth\cite{Ecodepth} † & 954M & 0.059 & 0.218 & 0.026 & 0.013 & 0.978 & 0.997 & 0.999\\
    DAR-B\cite{dar} † & 1.0B & 0.058 & 0.214 & 0.026 & \underline{0.013} & 0.980 & 0.997 & 0.999\\
    DAR-L\cite{dar} † & 2.0B & \textbf{0.056} & \textbf{0.205} & \textbf{0.024} & \textbf{0.011} & 0.982 & 0.998 & \textbf{1.000} \\
    \bottomrule
    ZoeDepth\cite{Zoedepth} * & 343M & 0.075 & 0.270 & 0.032 & 0.030 & 0.955 & 0.995 & 0.999\\
    DAv1\cite{DA1} * & 343M & \textbf{0.056} & \underline{0.206} & \textbf{0.024} & - &  \textbf{0.984} & \textbf{0.998} & \textbf{1.000} \\
    DAv2\cite{DA2} * & 343M & \textbf{0.056} & \underline{0.206} & \textbf{0.024} & - &  \textbf{0.984} & \textbf{0.998} & \textbf{1.000} \\
    \bottomrule
    DINOv3-B\cite{dinov3} & 120M & 0.072 & 0.252 & 0.031 & 0.026 & 0.964 & 0.997 & 0.999 \\
    \textbf{DINOv3-B+LFR} & 127M & 0.070 & 0.251 & 0.030 & 0.025 & 0.965 & 0.997 & 0.999 \\
    DINOv3-L\cite{dinov3} & 342M & 0.060 & 0.212 & 0.026 & 0.018 & 0.981 & 0.998 & \textbf{1.000} \\
    \textbf{DINOv3-L+LFR} & 350M & \underline{0.057} & \underline{0.206} & \underline{0.025} & 0.017 & \textbf{0.984} & \textbf{0.998} & \textbf{1.000} \\
  \bottomrule
  \end{tabular}
\end{table}

\begin{figure}[tb]
  \centering
  \includegraphics[height=9cm]{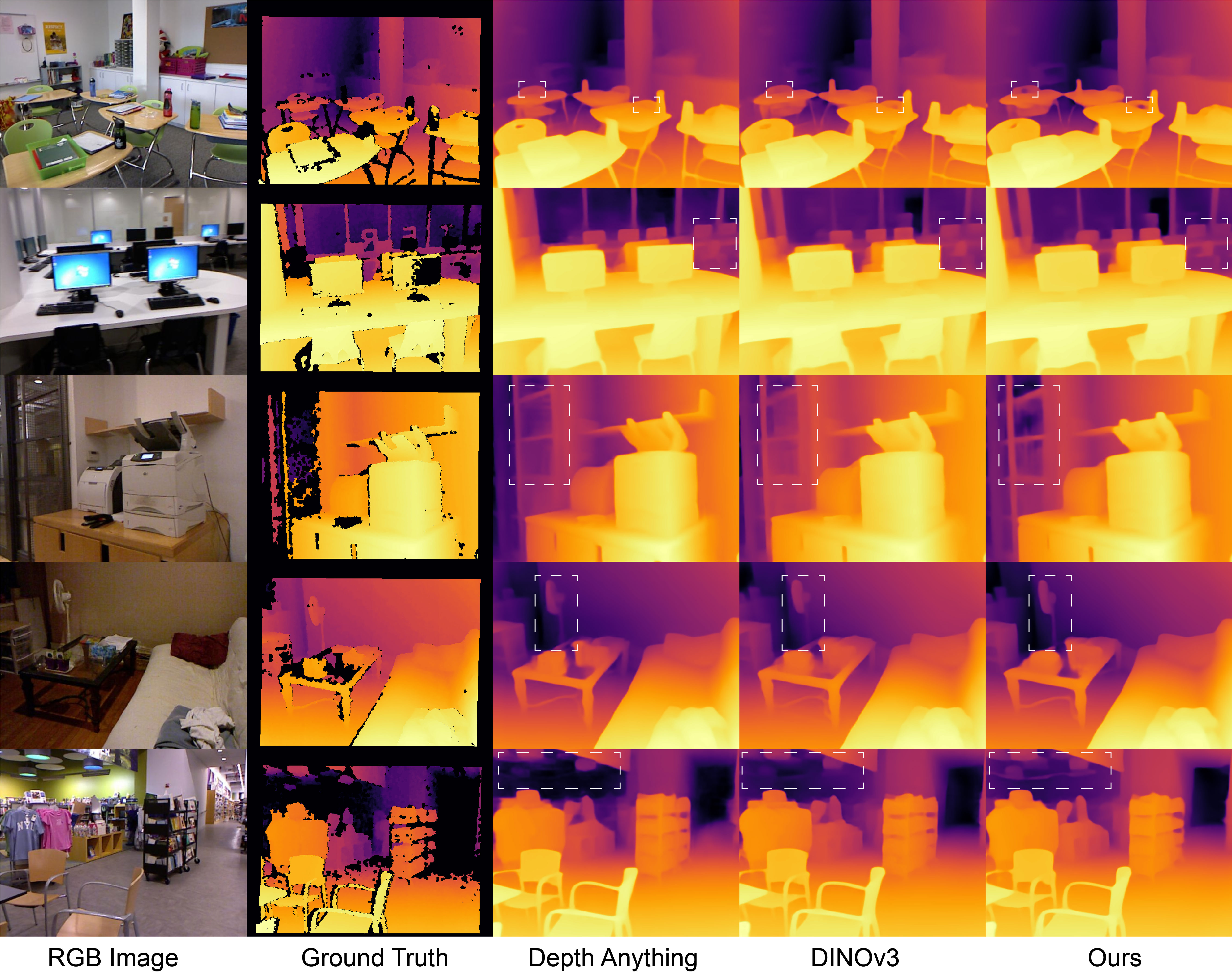}
  \caption{Qualitative depth predictions on NYU Depth v2. Brighter pixels indicate closer distances.
  }
  \label{fig:nyu}
\end{figure}

\begin{table}[tb]
  \caption{Comparison on KITTI. Notation follows Table 1.
  }
  \label{tab:kitti}
  \centering
  
  \begin{tabular}{@{}lllllllll@{}}
    \toprule
    Method & Size & AbsRel↓ & SqRel↓ & RMSElog↓ & RMSE↓ & \(\delta_1\)↑ & \(\delta_2\)↑ & \(\delta_3\)↑\\
    \midrule
   Eigenetal\cite{Eigen} & 45M & 0.203 & 1.517 & 0.282 & 6.307 & 0.702 & 0.898 & 0.967  \\
   DORN\cite{DORN} & 45M & 0.072 & 0.307 & 0.120 & 2.727 & 0.932 & 0.984 & 0.994 \\ 
   BTS\cite{BTS} & 20M & 0.059 & 0.241 & 0.096 & 2.756 & 0.956 & 0.993 & 0.998  \\
   AdaBins\cite{adabins} & 40M & 0.067 & 0.190 & 0.088 & 2.960 & 0.949 & 0.992 & 0.998  \\
   DPT\cite{DPT} & 343M & 0.060 & - & 0.092 & 2.573 & 0.959 & 0.995 & 0.996  \\
   P3Depth\cite{P3depth} & 45M & 0.071 & 0.270 & 0.103 & 2.842 & 0.953 & 0.993 & 0.998  \\
   NeWCRFs\cite{NeWCRF} & 270M & 0.052 & 0.155 & 0.079 & 2.129 & 0.974 & 0.997 & 0.999  \\
   PixelFormer\cite{PixelFormer} & 365M & 0.051 & 0.149 & 0.077 & 2.081 & 0.976 & 0.997 & 0.999  \\
   IEBins\cite{Iebins} & 273M & 0.050 & 0.142 & 0.075 & 2.011 & 0.978 & 0.998 & 0.999  \\
   MIM\cite{MIM} & 197M & 0.050 & 0.139 & 0.075 & 1.966 & 0.977 & 0.998 & 1.000  \\
   \bottomrule
   DDP\cite{DDP} † & 207M & 0.050 & 0.148 & 0.076 & 2.072 & 0.975 & 0.997 & 0.999  \\
   EcoDepth\cite{Ecodepth} † & 954M & 0.048 & 0.139 & 0.074 & 2.039 & 0.979 & 0.998 & 0.999  \\
   DAR-B\cite{dar} † & 1.0B & 0.046 & 0.114 & 0.069 & 1.832 & 0.985 & \textbf{0.999} & \textbf{1.000}  \\
   DAR-L\cite{dar} † & 2.0B & \underline{0.044} & 0.110 & \underline{0.067} & 1.799 & 0.986 & \textbf{0.999} & \textbf{1.000}  \\
   \bottomrule
   ZoeDepth\cite{Zoedepth} * & 343M & 0.054 & 0.189 & 0.083 & 2.440 & 0.970 & 0.996 & 0.999  \\
   DA v1\cite{DA1} * & 343M & 0.046 & - & 0.069 & 1.896 & 0.982 & 0.998 & \textbf{1.000}  \\
   DA v2\cite{DA2} * & 343M & 0.045 & - & \underline{0.067} & 1.861 & 0.983 & 0.998 & \textbf{1.000}  \\
   \bottomrule
   DINOv3-B\cite{dinov3} & 120M & 0.057 & 0.145 & 0.077 & 1.998 & 0.979 & 0.998 & \textbf{1.000}  \\
   \textbf{DINOv3-B+LFR} & 127M & 0.049 & 0.128 & 0.072 & 1.931 & 0.981 & 0.998 & \textbf{1.000}  \\
   DINOv3-L\cite{dinov3} & 342M & 0.046 & \underline{0.107} & \underline{0.067} & \underline{1.794} & \underline{0.986} & \textbf{0.999} & \textbf{1.000}  \\
   \textbf{DINOv3-L+LFR} & 350M & \textbf{0.043} & \textbf{0.099} & \textbf{0.063} & \textbf{1.711} & \textbf{0.988} & \textbf{0.999} & \textbf{1.000}  \\
  \bottomrule
  \end{tabular}
\end{table}

\begin{figure}[tb]
  \centering
  \includegraphics[height=4cm]{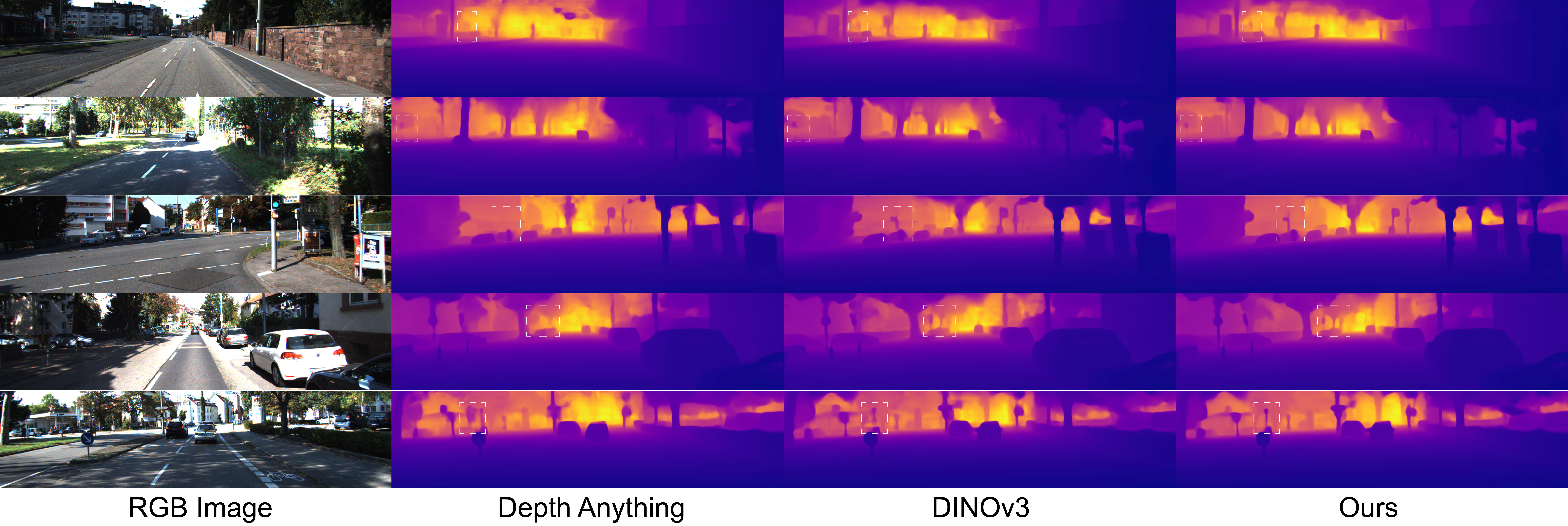}
  \caption{Qualitative depth predictions on KITTI. Brighter pixels indicate farther distances.
  }
  \label{fig:kitti}
\end{figure}

\section{Method}
\subsection{Preliminaries: DINOv3 Architecture}
\label{sec:line-numbering}
DINOv3 adopts a standard ViT architecture. Given an input image \(I\in\mathbb{R}^{H\times W\times3}\), a patch embedding layer \(P\left(\bullet\right)\) first partitions the image into non-overlapping patches and linearly projects each patch into a sequence of tokens: \(T_0 = P\left(I\right) \in \mathbb{R}^{N \times C}\), where \(N=\frac{H}{16}\times\frac{W}{16}\) denotes the number of tokens (i.e., the spatial resolution is reduced by a factor of 16) and C is the embedding dimension. This token sequence is then processed by a stack of \(\ L\) transformer encoder layers \(\left\{B_l\right\}_{l=1}^L\). Each layer consists of multi-head self-attention and feed-forward networks, progressively integrating global contextual information and refining high-level semantic representations.

For dense prediction tasks based on ViT, a common practice is to equidistantly sample features from four intermediate layers (e.g., layers 3, 6, 9, and 12 in ViT-B) to construct multi-scale representations that mimic the hierarchical feature pyramids of convolutional neural networks\cite{DPT}. Our analysis questions whether uniform sampling is optimal for transferring geometric knowledge.

\subsection{Feature Analysis of DINOv3}
\label{sec:line-numbering}
To understand how 3D geometric knowledge is distributed across layers, we conduct a systematic statistical analysis of DINOv3-Large \((L=24,\; C=1024)\). Specifically, we randomly sample 100 RGB images from the NYU Depth V2\cite{NYU}, feed them through DINOv3-L, and extract image tokens from all transformer layers, denoted as\(\left\{F_l\in\mathbb{R}^{N\times C}\right\}_{l=1}^L\). We then perform the following analyses:

\textbf{Representational Dissimilarity Across Samples.} For each layer, we compute the mean token 
representation per sample and calculate the pairwise Pearson correlation distance between samples, yielding a representational dissimilarity matrix \(\left\{{RDM}_l\in\mathbb{R}^{S\times S}\right\}_{l=1}^L\), where S is the number of samples. The average value of \({RDM}_l\) quantifies the inter-sample representational distance at layer \textit{l}.

\textbf{Representational Similarity to Depth Geometry.} We first construct a depth distance matrix \(\text{DDM} \in \mathbb{R}^{S \times S}\) by flattening each ground-truth depth map into a 1D vector and computing pairwise Pearson correlation distances. We then compute the Spearman correlation between \({RDM}_l)\) and \(\text{DDM}\) for each layer , measuring how well the representations capture variations in 3D geometry across samples.

\textbf{Depth Predictability via Linear Regression.} Each depth map is bilinearly downsampled to match the spatial resolution of the token grid and flattened into a depth vector. For each layer in each sample, we independently train a simple linear regression model that uses all channels of the image tokens as predictors and the corresponding depth vector as targets. The coefficient of determination \((R^2)\) between predicted and ground-truth depth serves as a layer-wise depth-predictability metric.

As illustrated in Fig. \ref{fig:attn}, our analysis reveals two key trends. First, features in the middle-to-deep layers tend to converge toward a shared representational space across samples, while the final layers shift toward more individualized representations. Second, depth predictability and the capacity to explain inter-sample variations in 3D geometry consistently increase with layer depth. Together, these trends indicate that deeper layers in DINOv3 encode richer and more explicit 3D information—suggesting they should be emphasized when transferring to MDE.

\subsection{Feature Analysis of DINOv3}
Motivated by the above analysis, we propose a lightweight, plug-and-play feature recombination module inserted between the ViT backbone and the prediction head. Without modifying the backbone, LFR centers on the final-layer features and adaptively selects a small number of complementary intermediate layers to fuse.

\textbf{Minimal-Similarity-Based Feature Selection.} Let \(F_L\) and \({Cls}_L\) denote the image tokens and class token of the final layer, which serve as the dominant features. We first compute the average cosine similarity between \({Cls}_L\) and the image tokens of each preceding layer. The K layers with the smallest similarity scores are selected as auxiliary layers (K=4 in our study), and their image tokens and class tokens are denoted as \(\left\{F_{ak},\ {Cls}_{ak}\right\}_{k=1}^K\). This minimal-similarity criterion ensures that the selected auxiliary features are semantically most complementary to the last-layer features, thereby maximizing information gain and minimizing redundancy. We also experiment with alternative selection strategies (e.g., maximal similarity, node degree, predicted-value-based selection), but find that minimal-similarity selection yields the best performance (Table \ref{tab:selection}).

\textbf{Feature Recombination.} For each selected auxiliary layer \(a_k\), we recombine its features with the dominant last-layer features using a lightweight adapter. Specifically, the auxiliary and dominant features are concatenated along the channel dimension to preserve their respective semantic information. The concatenated features are then passed through a compact adapter \(\mathrm{A}^k\), which consists of a linear down-projection layer, a non-linear activation function, and a linear up-projection layer, forming a bottleneck structure that facilitates feature fusion. A learnable scalar gate \(g^k\), modulated by a tanh function to constrain its range to \(\left[-1,1\right]\), controls the contribution of the fused features. The final recomposed features are obtained via a residual connection:

\begin{equation}
F_{rk}=Tan{\left(g^k\right)\bullet}A^k\left(\left[F_{ak};F_L\right]\right)+F_L
\end{equation}

\begin{equation}
{Cls}_{rk}=Tan{\left(g^k\right)\bullet}A^k\left(\left[{Cls}_{ak};{Cls}_L\right]\right)+{Cls}_L
\end{equation}

The recomposed features \(\left\{F_{rk}\right\}_{k=1}^K\) and \(\left\{{Cls}_{rk}\right\}_{k=1}^K\) are subsequently used for depth regression.

\textbf{Prediction Head.} We adopt a DPT-style decoder\cite{DPT} for depth regression but modify the aggregation strategy to exploit the recomposed features. In the original DPT, multi-scale features are assembled into a feature pyramid, and only the final layer's features are used for prediction. In contrast, we perform depth regression independently for each recomposed feature level. Specifically, after constructing feature pyramids from \(\left\{{Cls}_{rk}\right\}_{k=1}^K\}_{i=1}^4\), we obtain per-level depth predictions. The final depth map is computed as a weighted sum of these per-level predictions, where the weights are derived from the recomposed class tokens:
\begin{equation}
w^k=Liner\left({Cls}_{rk}\right),\ {w^k}_{k=1}^K=\mathrm{softmax}({w^k}_{k=1}^K)
\end{equation}

and the final prediction is \(\widetilde{D}=\sum_{k=1}^{K}w^k\cdot{\widetilde{D}}_k\), with \({\widetilde{D}}_k\) denoting the depth prediction from level \textit{k}. This aggregation strategy allows each recomposed feature, which already encodes rich 3D knowledge centered on the last layer, to contribute according to its confidence, enhancing prediction robustness.

\textbf{Loss Functions.} We employ two complementary loss functions to supervise training. The first is the scale-invariant log loss\cite{Eigen}, commonly used in MDE:
\begin{equation}
\mathcal{L}_{slog}=\sqrt{\frac{1}{N}\sum_{n=1}^{N}g_n^2-\frac{\lambda}{N^2}\left(\sum_{n=1}^{N}g_n\right)}
\end{equation}

where \(g_n=\log{{\widetilde{d}}_n-\log{d_n}}\), \(d_n\) and \({\widetilde{d}}_n\) are ground-truth and predicted depth values, N is the number of valid pixels, and \(\lambda\) is a balancing parameter (set to 0.85 following prior work).
The second is a hierarchical normalization loss\cite{hn} that integrates multi-scale depth normalization to balance global coherence and local detail. The depth map is partitioned into patches at multiple scales \(M\in{1,4,8}\). For each patch \textit{u} at scale \textit{M}, both predicted and ground-truth depth are normalized within the patch. The per-pixel loss is then averaged over all patches containing that pixel:
\begin{equation}
\mathcal{L}_{HN}^n={\frac{1}{\left|U_n\right|}\sum_{u\in U_n}\left|\mathcal{N}_u\left(d_n\right)-\mathcal{N}_u\left({\widetilde{d}}_n\right)\right|}
\end{equation}

where \(U_n\) denotes the set of patches containing pixel n, and \(\mathcal{N}_u\) denotes normalization within patch u. The final hierarchical loss is averaged over all valid pixels:
\begin{equation}
\mathcal{L}_{HN}=\frac{1}{N}\sum_{n=1}^{N}L_{HN}^n.
\end{equation}

This loss balances global coherence and local detail, and empirically helps elicit explicit 3D structures from the recomposed features.The final training objective is \(\mathcal{L}=\mathcal{L}_{\mathrm{slog}}+\mathcal{L}_{\mathrm{HN}}\)

\begin{table}[tb]
  \caption{Zero-shot generalization on SUN RGB-D. Models trained on NYU Depth v2 are directly evaluated on SUN RGB-D without fine-tuning.
  }
  \label{tab:sun}
  \centering
  \setlength{\tabcolsep}{4pt}
  \begin{tabular}{@{}lllll@{}}
    \toprule
    Method & AbsRel↓ & RMSE↓ & log10↓ & \(\delta_1\)↑ \\
    \midrule
  BTS\cite{BTS} & 0.143 & 0.421 & 0.061 & 0.805 \\
  AdaBins\cite{adabins} & 0.159 & 0.476 & - & 0.768 \\
  NeWCRFs\cite{NeWCRF} & 0.15 & 0.429 & - & 0.799 \\
  VPD\cite{VPD} & 0.121 & 0.355 & 0.045 & 0.861 \\
  EcoDepth\cite{Ecodepth} & 0.112 & 0.319 & - & 0.885 \\
  DA v1\cite{DA1} & 0.119 & 0.346 & 0.043 & 0.864 \\
  DAR-L\cite{dar} & 0.112 & 0.319 & 0.040 & 0.885 \\
  DINOv3-L\cite{dinov3} & 0.101 & 0.312 & 0.046 & 0.895 \\
  \bottomrule
  \textbf{DINOv3-L+LDA} & \textbf{0.099} & 0.\textbf{300} & \textbf{0.044} & \textbf{0.907} \\
  \bottomrule
  \end{tabular}
\end{table}

\begin{table}[tb]
  \caption{Zero-shot generalization on Argoverse. Models trained on KITTI are directly evaluated on Argoverse without fine-tuning.
  }
  \label{tab:argo}
  \centering
  \setlength{\tabcolsep}{4pt}
  \begin{tabular}{@{}lllll@{}}
    \toprule
    Method & AbsRel↓ & RMSE↓ & SIlog↓ & \(\delta_1\)↑ \\
    \midrule
  BTS\cite{BTS} & 0.307 & 15.98 & 0.518 & 0.307 \\
  AdaBins\cite{adabins} & 0.383 & 17.07 & 0.523 & 0.383 \\
  P3Depth\cite{P3depth} & 0.277 & 17.97 & 0.441 & 0.277 \\
  NeWCRF\cite{NeWCRF} & 0.311 & 15.75 & 0.468 & 0.311 \\
  iDisc\cite{idisc} & 0.560 & 12.18 & 0.334 & 0.560 \\
  DINOv3-L\cite{dinov3} & 0.219 & 11.13 & 0.317 & 0.677 \\
  \bottomrule
  \textbf{DINOv3-L+LFR} & \textbf{0.204} & \textbf{10.830} & \textbf{0.309} & \textbf{0.730} \\
  \bottomrule
  \end{tabular}
\end{table}

\begin{table}[tb]
  \caption{Ablation studies. Starting from the baseline (DINOv3 + DPT), we progressively add each component of our method.
  }
  \label{tab:ablation}
  \centering
  \setlength{\tabcolsep}{2pt}
  \begin{tabular}{@{}lllllllll@{}}
    \toprule
    Last layer & Recombine & HN loss & Multi-level 
 & AbsRel↓ & SqRel↓ & RMSE↓ & SIlog↓ & \(\delta_1\)↑ \\
    \midrule
  $\times$  & $\times$  & $\times$  & $\times$  & 0.046 & 0.107 & 1.794 & 0.062 & 0.986 \\
  $\checkmark$ & $\times$  & $\times$  & $\times$  & 0.045 & 0.105 & 1.743 & 0.061 & 0.987 \\
  $\checkmark$ & $\checkmark$ & $\times$  & $\times$  & 0.043 & 0.099 & 1.721 & 0.060 & 0.987 \\
  $\checkmark$ & $\checkmark$ & $\checkmark$ & $\times$  & 0.043 & 0.099 & 1.719 & 0.059 & 0.987 \\
  $\checkmark$  & $\checkmark$  & $\checkmark$  & $\checkmark$  & 0.043 & 0.099 & 1.711 & 0.059 & 0.988 \\
  \bottomrule
  \end{tabular}
\end{table}

\begin{table}[tb]
  \caption{Comparison of different auxiliary layer selection strategies. 'Scores': layers are selected based on a score predicted from their class tokens. 'Node degree': layers with the smallest average similarity between class and image tokens are selected. 'Maximal-similarity': layers with the highest average similarity to the last-layer class token are selected. 'Minimal-similarity': our used strategy selecting layers with the lowest similarity to the last-layer class token.
  }
  \label{tab:selection}
  \centering
  \setlength{\tabcolsep}{3pt}
  \begin{tabular}{@{}lllllllll@{}}
    \toprule
    Method & AbsRel↓ & SqRel↓ & RMSElog↓ & RMSE↓ & SIlog↓ & \(\delta_1\)↑ \\
    \midrule
  Scores & 0.044 & 0.101 & 0.064 & 1.738 & 0.060 & 0.987 \\
  Node degree & 0.044 & 0.100 & 0.064 & 1.721 & 0.060 & 0.987 \\
  Maximal-similarity & 0.044 & 0.100 & 0.064 & 1.723 & 0.060 & 0.987 \\
  Minimal-similarity & 0.043 & 0.099 & 0.063 & 1.711 & 0.059 & 0.988 \\
  \bottomrule
  \end{tabular}
\end{table}

\section{Experiments}
\subsection{Datasets}
We evaluate the proposed method on two standard MDE benchmarks: 

\textbf{NYU-Depth v2}\cite{NYU}. An indoor dataset with 464 scenes; we follow standard practice and use 24,231 training images and 654 test images at \(640\times480\)resolution with depth range 0-10m. \textbf{KITTI}\cite{kitti}. An outdoor driving dataset with stereo images and Velodyne LiDAR scans. We use images resized to \(1241\times376\) and adopt the Eigen split\cite{Eigen} (23,158 training and 652 test samples), depth range 0-80m.

For zero-shot evaluation, we employ two additional datasets: \textbf{SUN RGB-D}\cite{Sunrgb-d} for indoor generalization, depth range 0-10m (resized to \(640\times480\), official test set of 5,050 samples).\textbf{Argoverse}\cite{Argo} for outdoor transfer, depth range 0-150m (images resized to \(1241\times376\); follow evaluation protocol in \cite{idisc} to select 476 test samples).

\subsection{Implementation Details}
Our implementation uses PyTorch and training is performed on four NVIDIA A100 GPUs. We train for 30 epochs using AdamW\cite{adam} (\(\beta_1=0.9,\beta_2=0.999\)) with batch size 8. Learning rates are \(({10}^{-5}\) for the backbone and \(({10}^{-4}\) for adapters and the prediction head. We apply a linear warmup over the first 10\% of iterations followed by cosine annealing; the backbone is frozen after 20 epochs. Data augmentation includes random rotation, scaling, and photometric jittering (brightness, gamma, saturation, hue). Training one epoch takes 7 minutes; a single inference on a \(640\times480\) image costs approximately 680 GFLOPs.

\subsection{Comparison on NYU-Depth v2}
Table \ref{tab:nyu} reports quantitative results on the NYU-Depth v2 dataset. A vanilla DINOv3\cite{dinov3} backbone with a standard DPT decoder already attains competitive performance; augmenting it with LFR yields consistent improvements across all standard metrics. Compared to prior SOTA methods—which often rely on multi-step generative frameworks (diffusion/autoregressive) or extensive relative-depth pretraining—our approach achieves superior or comparable accuracy while avoiding expensive iterative inference or massive pretraining. Qualitative examples (Fig. \ref{fig:nyu}) show improved boundary fidelity and preservation of fine geometric structures.
\subsection{Comparison on KITTI}
Table. \ref{tab:kitti} presents results on the KITTI dataset. Inserting our LFR module into DINOv3 yields substantial improvements over the baseline and outperforms all previous methods, including those based on generative modeling or large-scale pre-training. The gains are particularly pronounced for distant small objects (Fig. \ref{fig:kitti}), indicating enhanced geometric understanding.

\begin{figure}[tb]
  \centering
  \includegraphics[height=6cm]{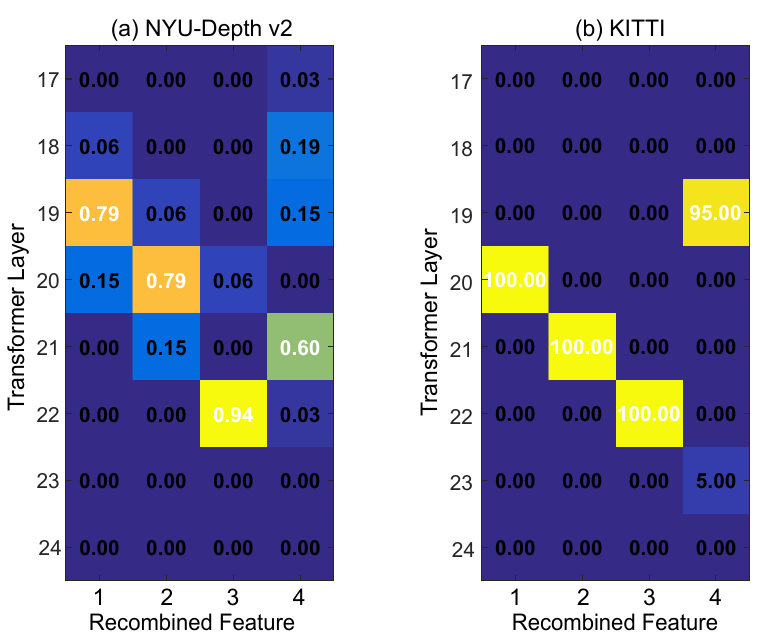}
  \caption{Distribution of selected auxiliary layers. Statistics are computed over 100 random samples per dataset. Numbers indicate the proportion of times each layer is selected; each column sums to 1.
  }
  \label{fig:sele}
\end{figure}

\begin{figure}[tb]
  \centering
  \includegraphics[height=4cm]{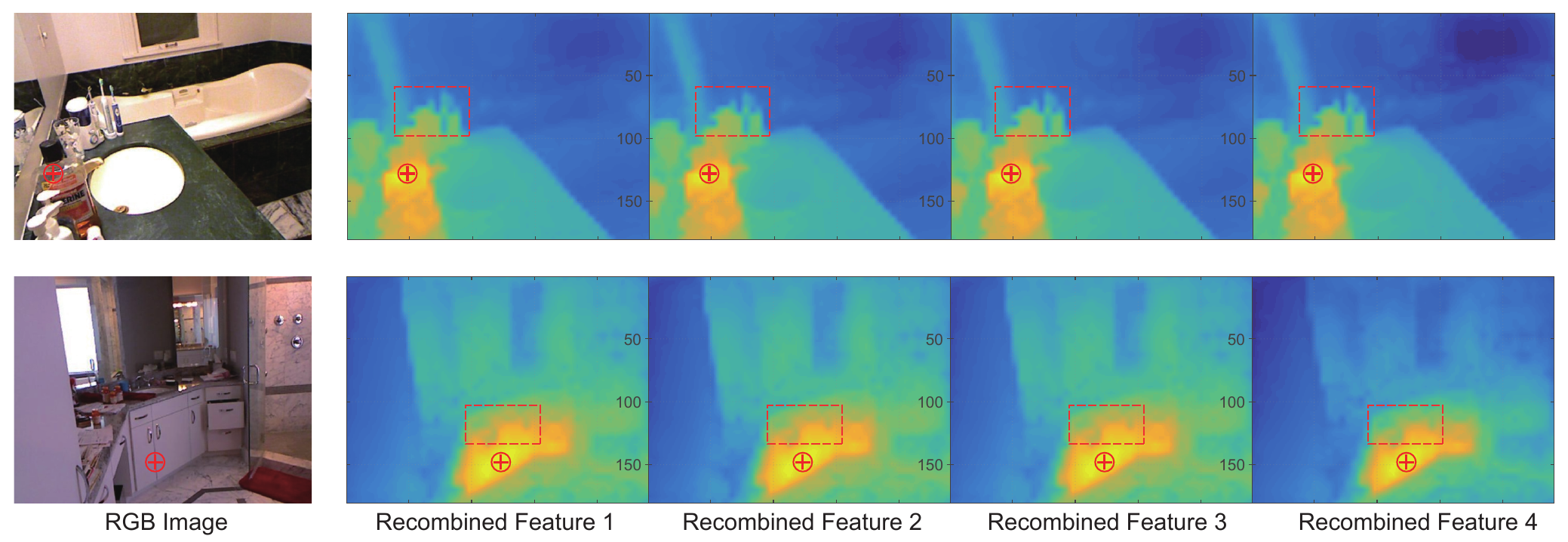}
  \caption{Attention maps of recomposed features. For each recomposed feature level, we visualize the cosine similarity between the anchor token and all other tokens. Brighter regions indicate higher similarity.
  }
  \label{fig:attn_adap}
\end{figure}

\subsection{Zero-Shot Generalization}
To assess cross-dataset generalization, we directly transfer models trained on NYU-Depth v2 to SUN RGB-D (Table \ref{tab:sun}) and models trained on KITTI to Argoverse (Table. \ref{tab:argo}), without any fine-tuning. Our method consistently outperforms prior approaches, demonstrating superior zero-shot transferability. This confirms that our feature recombination strategy effectively extracts and reinforces generic 3D geometric priors embedded in the VFMs, enabling robust adaptation to unseen environments.
\subsection{Ablation Studies}
Table. \ref{tab:ablation} ablates the key components of our method. Starting from a strong baseline (DINOv3 + DPT), we evaluate the following variants:
\begin{itemize}
    \item[•] Last-layer only: Replicating the last-layer features four times and feeding them into DPT already improves performance over the uniform-sampling baseline, supporting our finding that deeper layers encode rich multi-scale information sufficient for dense prediction.
    \item[•] + Recombination module: Adding our feature recombination module brings a significant further gain, validating that the selected intermediate layers provide complementary information.
    \item[•] + HN loss: Incorporating the hierarchical normalization loss yields additional improvement, indicating that multi-scale normalization helps elicit explicit 3D knowledge.
    \item[•] + Multi-level weighted prediction: Finally, using weighted aggregation of per-level predictions further boosts accuracy, demonstrating that each recomposed feature contributes useful cues.
\end{itemize}

\subsection{Analysis of Layer Selection}
Table \ref{tab:selection} compares different strategies for selecting auxiliary layers. All strategies outperform the last-layer-only baseline, confirming that intermediate layers provide complementary information. Among them, minimal-similarity selection achieves the best results, likely because it maximizes semantic complementarity and reduces redundancy.

Fig. \ref{fig:sele} visualizes the distribution of selected layer indices (over 100 random samples per dataset). The selected auxiliary layers predominantly lie in the middle-to-deep range (layers 17–22), which, as shown in Fig. X, encode substantial 3D knowledge but exhibit lower inter-sample variation. This suggests that injecting shared representations from these layers into the more individualized final-layer space improves prediction stability. Interestingly, the selected layers are more diverse for indoor scenes (NYU Depth v2\cite{NYU}) than for outdoor scenes (KITTI\cite{kitti}), possibly reflecting the higher complexity and variability of indoor environments.

\subsection{Attention Visualization}
Fig. \ref{fig:attn_adap} visualizes attention maps derived from the recomposed features. All maps exhibit similar spatial patterns, accurately highlighting regions corresponding to query points, with differences only in fine details. This indicates that each recomposed feature retains the strong localization capability of the last layer while offering subtle variations in visual bias, which collectively enhance robustness across diverse scenes.

\section{Conclusion}
In this work, we focused on effectively transferring the powerful DINOv3 to MDE. Recognizing that DINOv3 inherently possesses strong 3D understanding capabilities, we first conducted a comprehensive layer-wise feature analysis, revealing that deeper layers encode richer and more critical 3D geometric knowledge. Motivated by this insight, we proposed a simple yet effective method that recombines the last-layer features with carefully selected intermediate-layer features, thereby further unlocking DINOv3's inherent strengths for MDE. Our LFR module is lightweight, plug-and-play, and requires no modification to the backbone. Extensive experiments on multiple benchmarks demonstrate that our approach consistently outperforms SOTAs, achieving superior accuracy and strong zero-shot generalization. Beyond providing an advanced tool for MDE, our work offers theoretical guidance for transferring VFMs to downstream tasks by revealing how 3D knowledge is distributed across layers. In future work, we plan to extend this method to other 3D understanding tasks and dense prediction problems, further exploring the potential of LFR.

\clearpage  

\section*{Acknowledgements}
Please insert your acknowledgments here.

%
%
\bibliographystyle{splncs04}
\bibliography{main}
\end{document}